\documentclass[10pt,twocolumn,letterpaper]{article}
\usepackage{cvpr}              % To produce the CAMERA-READY version
\usepackage{epsfig}
\usepackage{amsmath}
\usepackage{amssymb}
\usepackage{graphicx}
\usepackage{tikz}
\usepackage{comment}
\usepackage{color}
\usepackage{pifont}
\usepackage{epsfig}
\usepackage{graphics}
\usepackage{multirow}
\usepackage{algorithm}
\usepackage{algorithmicx}
\usepackage{subcaption}
\usepackage{sidecap}
\usepackage{wrapfig}
\usepackage{enumitem}
\usepackage{colortbl}
\usepackage{arydshln,xcolor,array}
\usepackage{multirow}
\usepackage{esvect}
\usepackage{xcolor}

\usepackage{graphicx}
\usepackage{booktabs}
\newcommand*\colourcheck[1]{%
  \expandafter\newcommand\csname #1check\endcsname{\textcolor{#1}{\ding{51}}}%
}
\newcommand*\colourcross[1]{%
  \expandafter\newcommand\csname #1cross\endcsname{\textcolor{#1}{\ding{55}}}%
}
\colourcheck{green}
\colourcross{red}
\let\origtau\tau % sav
\renewcommand{\tau}{\scalebox{1.2}{$\origtau$}}
\definecolor{cvprblue}{rgb}{0.21,0.49,0.74}
\usepackage[pagebackref,breaklinks,colorlinks,citecolor=cvprblue]{hyperref}

\def\paperID{142} % *** Enter the Paper ID here
\def\confName{3DV\xspace}
\def\confYear{2025\xspace}
\newcommand{\model}[0]{FORCE~}
\newcommand{\vect}[1]{\boldsymbol{#1}}
\newcommand{\mat}[1]{\mathbf{#1}}
\newcommand{\set}[1]{\mathcal{#1}}
\newcommand{\removespace}{\hspace{-\fontdimen2\font plus -\fontdimen3\font minus -\fontdimen4\font}}

\newcommand{\contact}{\vec{c}}

\newcommand{\resistance}{\mat{R}} 
 
\newcommand{\force}{\mat{F}} 
\newcommand{\occupancy}{\mat{c}}

\newcommand{\acceleration}{\vect{a}}

\newcommand{\humanphase}{\mat{\Phi}}

\newcommand{\poseposition}{\vect{j}^{p}_i}
\newcommand{\posevelocity}{\vect{j}^{v}_{i}}
\newcommand{\poserotation}{\vect{j}^{r}_{i}}

\newcommand{\trajpos}{\vect{t}^{p}_{i}}
\newcommand{\trajdir}{\vect{t}^{d}_{i}}

\newcommand{\trajact}{\vect{t}^{a}_{i}}

\newcommand{\goalpos}{\vect{o}^{p}_{i}}
\newcommand{\goaldir}{\vect{o}^{d}_{i}}
\newcommand{\goalact}{\vect{o}^{a}_{i}}

\newcommand{\pose}{\mat{J}}
\newcommand{\traj}{\mat{T}}
\newcommand{\goal}{\mat{O}}
\newcommand{\geom}{\mat{G}}

\newcommand{\futureposeposition}{\vect{\tilde{j}}^{p}_{i+1}}

\newcommand{\goaltraj}{\mat{\tilde{T}}}

\newcommand{\latent}{\vect{z}}

\newcommand{\myparagraph}[1]{\vspace{1pt}\noindent{\bf #1}}

\usepackage{graphicx} % For including images
\usepackage{caption} % For handling captions
\usepackage[utf8]{inputenc} % Ensure proper encoding

\title{FORCE: Physics-aware Human-object Interaction}
\begin{document}

% Manually define authors
\author{
    Xiaohan Zhang$^{1,2}$, Bharat Lal Bhatnagar$^{3}$, Sebastian Starke$^{3}$, 
    Ilya Petrov$^{1,2}$,\\ Vladimir Guzov$^{1,2}$, Helisa Dhamo$^{4}$, Eduardo P\'erez-Pellitero$^{4}$, Gerard Pons-Moll$^{1,2}$ \\
}

% Customize maketitle to include affiliations above the image
\makeatletter
\let\@oldmaketitle\@maketitle % Store the original \@maketitle
\renewcommand{\@maketitle}{%
    \@oldmaketitle % Keep the default title and authors
    \vspace{-1.5em} % Space after authors
    \begin{center}
        \normalsize
        $^1$Tübingen AI Center, University of Tübingen \\
        $^2$Max Planck Institute for Informatics, Saarland Informatics Campus \\
        $^3$Meta Reality Labs Research\\
        $^4$Huawei Noah's Ark Lab \\
        \vspace{2em} % Space before the image
        \includegraphics[width=.9\textwidth]{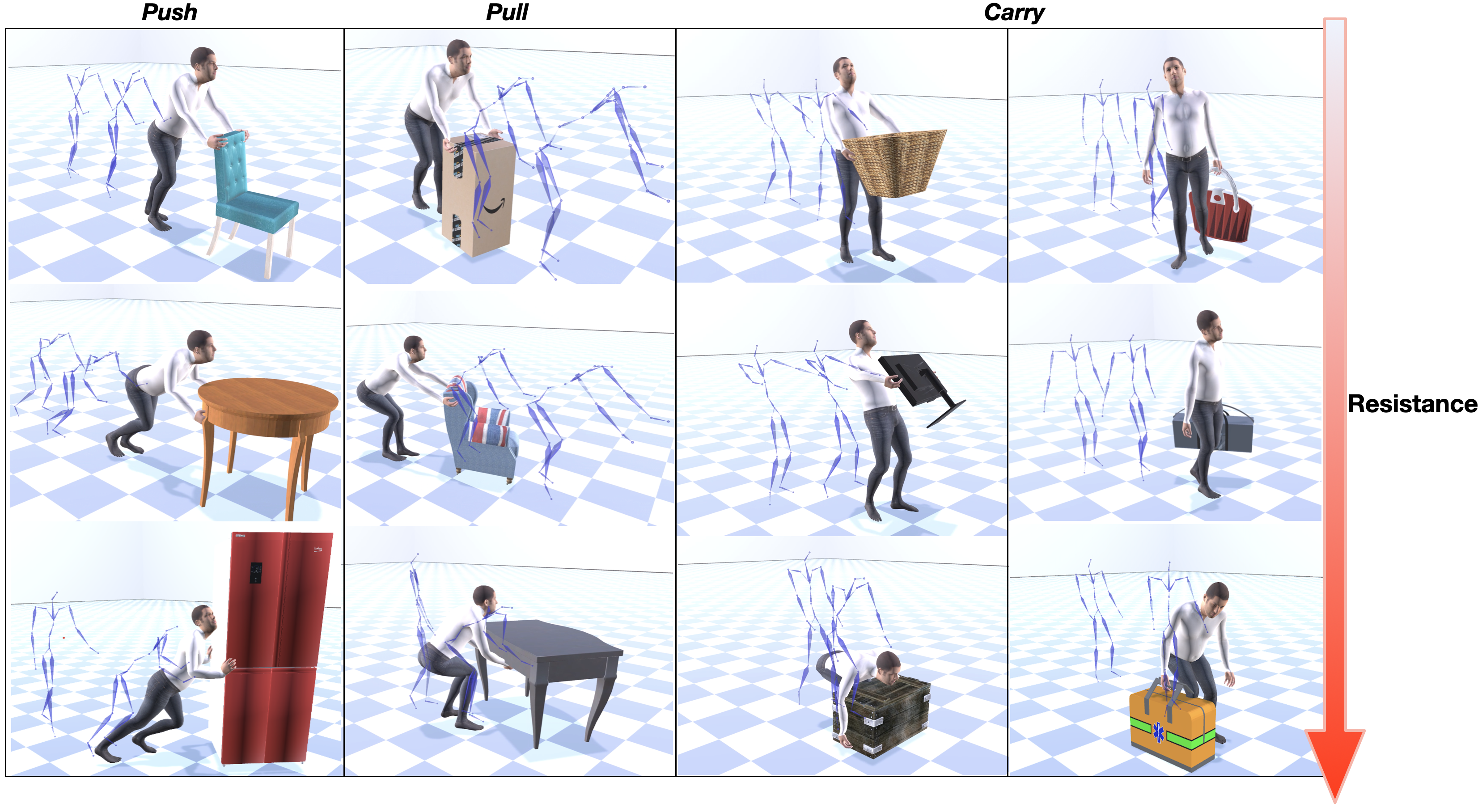} % Adjust path and width as needed
        % \vspace{1em} % Space after the image
    \end{center}
    \refstepcounter{figure}
    \centering \textbf{Figure~\thefigure}: \textit{\model~synthesizes diverse, nuanced human-object interactions by modeling the primary physical attributes.}
    \label{fig:teaser}
    \bigskip
}
\makeatother

\maketitle

\begin{abstract}

Interactions between human and objects are influenced not only by the object's pose and shape, but also by physical attributes such as object mass and surface friction. They introduce important motion nuances that are essential for diversity and realism. Despite advancements in recent human-object interaction methods, this aspect has been overlooked.
Generating nuanced human motion presents two challenges. First, it is non-trivial to learn from multi-modal human and object information derived from both the physical and non-physical attributes. Second, there exists no dataset capturing nuanced human interactions with objects of varying physical properties, hampering model development.
This work addresses the gap by introducing the FORCE model, an approach for synthesizing diverse, nuanced human-object interactions by modeling physical attributes. Our key insight is that human motion is dictated by the interrelation between the force exerted by the human and the perceived resistance. Guided by a novel intuitive physics encoding, the model captures the interplay between human force and resistance. Experiments also demonstrate incorporating human force facilitates learning multi-class motion.
Accompanying our model, we contribute a dataset, which features diverse, different-styled motion through interactions with varying resistances. Our code, dataset, and models will be released at \url{https://virtualhumans.mpi-inf.mpg.de/force/}.

\vspace{-10pt}
\end{abstract} 

\section{Introduction}

Synthesizing realistic motion of human-object interactions (HOI) presents a significant challenge. The complexity lies in the intricate interplay between humans and objects. While previous work has primarily addressed fundamental aspects of interactions, such as object shape and location, they overlook crucial physical attributes like weight, friction, and human force~\cite{nsm,samp,zhang2022couch,taheri2021goal}. Consequently, existing models struggle with distinguishing between carrying an empty suitcase and a full one, or determining if the interaction is feasible. These nuances are essential for diversity and realism. This work aims to fill this gap by leveraging the physical attributes to synthesize nuanced human motion in these scenarios.

As a parallel paradigm, physics-based methods using reinforcement learning coupled with imitation learning~\cite{peng2021amp,2022-TOG-ASE, Dou23csae,xie2023box} demonstrate remarkable generalization for different scenarios. However, they face challenges in providing direct fine-grained control, for example, the switching between left-handed carrying to using both hands. Additionally, these methods are computationally demanding, because distinctive policies need to be trained with tailored reward functions for navigation and interaction tasks.

Kinematic methods for human motion synthesis are known for their flexibility for fine-grained spatial or temporal motion control. They are scalable to train and easy to deploy. However, earlier approaches either disregard the surrounding environment~\cite{martinez,quaternet,Mao2019ICCV,aist} or focus solely on static objects~\cite{samp,taheri2021goal,wu2022saga,zhang2022couch,mir23origin,zhang2023roam}. Closest to our goal are methods~\cite{nsm,samp,holden2020,li2023OMOMO} that model object shapes, but overlook physical attributes of interaction. In reality, humans adapt their motion based on perceived resistance and applied force during an object interaction~\cite{intuitive1983,DuanDFT22}.
Take the \textit{pushing} action shown in Figure~\ref{fig:teaser} (left-most column). When pushing around a heavier object, the human applies a greater force, naturally adjusting posture, shifting the center of mass~\cite{tripathi2023ipman,com}, and leaning forward to push against the friction. If the resistance exceeds the applied force, the object would not move and the human would have to give up the interaction because it is not feasible. Synthesizing such nuanced motion calls for a method that generalizes well to these physical attributes of the interaction.

% challenge
Developing such a method presents multiple challenges. Firstly, it is non-trivial to reason about multi-modal human-object information, including different actions, object poses, and crucial physical attributes. The increased complexity makes it challenging to disambiguate similar human poses, leading to motion that lacks diversity and nuanced details. Secondly, when determining the feasibility of the interaction, resistance is not the sole factor. It also depends on how the human interacts. For instance, a human can more effectively carry a heavier object with two hands than with one. It is shown that naively conditioning on the resistance generates sub-optimal results (see Section~\ref{sec:experiments}). Additionally, there is no dataset capturing diverse daily interactions under varying physical conditions. The lack of such data hinders model development and evaluation. Challenges present even when collecting such data, when addressing issues such as object occlusion.

% solution / insight expanded
To address these challenges, we introduce~\model\removespace, a method for synthesizing human-object interactions that focuses on the nuanced details in human motion. Our method is founded on a pivotal insight: Human motion is dictated by the interrelation between the force exerted by the human and the perceived resistance. Guided by a novel intuitive physics encoding derived from these crucial attributes, our model is able to synthesize a diverse spectrum of interactions. For example, under the category of ``carrying'', our model can plausibly generate motions such as carrying an object, carrying an object followed by a need to drop it, or attempting to carry but encountering failure. Additionally, we enable interactive control at run-time, the style of motion can be manipulated not only by varying the resistance of the object, but also by the desired action and contact mode (left, right, two-handed).

Complementing our \model model, we present a dataset featuring diverse motion nuances through interactions involving 3-6 levels of resistance. We adopt a customized hybrid tracker comprising of 4 Kinect RGB-D cameras~\cite{kinect} paired with 17 Inertial Measurement Units (IMUs)~\cite{xsens}. Our novel dataset contains 450 motion sequences (192k frames) of pervasive interactions of carrying, pushing and pulling objects. For each frame, we provide high-quality human and object poses. This dataset can serve as a benchmark for various human-object interaction tasks.
The overall contributions of our work are: 
\begin{enumerate}
    \item We introduce \model\removespace, the first kinematic method to synthesize human-object interaction by modeling physical attributes such as resistance and the applied human force. It achieves state-of-the-art results quantitatively and qualitatively.
    \item To enable the synthesis of diverse, nuanced human motion, we propose a novel intuitive physics encoding. 
    \item We present a dataset that accurately captures the daily interactions of pushing, pulling and carrying objects. It features diverse, different-styled interaction motions with varying resistances. 
    \item We will release our code, dataset, and models to stimulate further research.
\end{enumerate}

\section{Related Work}
\myparagraph{Human Motion Synthesis.}
Human motion synthesis has been a long-standing challenge in computer vision, evolving from early non-contextual approaches~\cite{HenterAB20, AlexandersonNBH23, PerezHBHOA21, DBLP:journals/corr/abs-2306-00416,zhang2024tedi,chen2023humanmac,kim2022flame,DBLP:conf/aaai/LeeML23,barquero2023belfusion}, to more sophisticated methods. Recent advancements include text-to-motion~\cite{tevet2023human, shafir2023human,zhang2022motiondiffuse,zhang2023remodiffuse,barquero2024seamless,chen2023executing,dabral2022mofusion,DBLP:conf/iccv/KongGLMW23,DBLP:conf/aaai/HoangGGM24,liang2024intergen,humantomato,ma2024richcat,huang2024como}, motion-to-text~\cite{zhang2023motiongpt,jiang2024motiongpt}, motion synthesis with spatial~\cite{karunratanakul2023gmd,xie2024omnicontrol,DBLP:journals/corr/abs-2311-17135,diomataris2024wandr} or temporal control~\cite{petrovich24stmc}, audio to motion~\cite{Mehta2024fake,Chhatre2024CVPR}. Other directions explore human-to-human interaction~\cite{ghosh2024remos,DBLP:journals/corr/abs-2312-08983} and learning 3D human motions from images and videos~\cite{xu2021d3dhoi,yang2023lemon,ye2023affordance}.

In the context of human-object interactions, research has progressed from predicting static affordances in 3D scenes~\cite{3d-affordance,qi2018human, Zhang2020CVPR,PLACE:3DV:2020,PROX:2019, Zhao:ECCV:2022, Narrator,Hassan:CVPR:2021,li2024genzi} to dynamic human motions. However, most existing work focuses on human interactions with static scenes~\cite{wang2022humanise, huang2023diffusion, lee2023lama,mir23origin,PROX:2019,wang2022sceneaware, wang2020,cong2024laserhuman,li2024egogen}, while some studies have attempted to enhance motion generation quality by concentrating on static object interactions, such as sitting and lying on furniture~\cite{samp, zhang2023roam, zhang2022couch, Pi2023ICCV, pan2023synthesizing, Zhao:ICCV:2023,yi2024tesmo,kulkarni2023nifty}.\\
A parallel paradigm of research explores reinforcement learning (RL) for object manipulation~\cite{christen2022dgrasp,braun2023physically,dfbgrasp2024braun}, carrying objects~\cite{hassan2023,wan2022learn, merel2020catchcarry,xie2023box}, and performing locomotion~\cite{motionvae, rempeluo2023tracepace}. RL has also been applied to interactions with static objects~\cite{pan2023synthesizing, xiao2023unified} sports movements~\cite{wang2023physhoi} and text to motion~\cite{cui2024anyskill}. Another avenue of exploration centers around full-body grasping synthesis~\cite{taheri2021goal, wu2022saga,araujo2023circle,tendulkar2022flex, DBLP:conf/wacv/Li0L024} and simultaneous synthesis of human and object motions~\cite{ghosh2022imos,Li2023task,liu2024geneoh,MACS2024}. Specialized efforts are dedicated to synthesizing dexterous hand-object interactions using object motion~\cite{taheri2023grip,manipnet, zhou2022toch, petrov2023popup}. 

While these approaches have made significant strides, they primarily deal with static object interactions or specific types of movements. Our work addresses more challenging scenarios involving humans interacting and moving objects.

\myparagraph{Interactions with moving objects.}
Recent datasets have stimulated development in human-object interaction synthesis, capturing single-person~\cite{jiang2023chairs,zhao2024imhoi,huang2022intercap,bhatnagar22behave,li2023OMOMO,jiang2024scaling,kim2024parahome} and multi-person~\cite{zhang2024core4d,zhang2024hoi} with dynamic objects. Among the current methods for synthesizing dynamic human-object interactions, the Neural State Machine~\cite{nsm} can model both static and dynamic interactions. Diffusion models~\cite{diller2023cghoi, xu2023interdiff,peng2023hoi,DBLP:journals/corr/abs-2403-11208,xu2024interdreamer,yang2024fhoi} 
are applied to synthesize short-term human-object interactions with a fixed set of objects~\cite{bhatnagar22behave}, often assuming stable contact between humans and objects. More recent work~\cite{li2023OMOMO,li2024chois} conditions on object trajectories or waypoints to predict full-body object manipulation.

However, these previous datasets and methods overlook how human motion is dictated by the physical attributes of the interaction, such as mass or friction. Our work addresses this gap by capturing a dataset that focuses on interactions with objects of varying resistance, including pushing, pulling, and carrying. We emphasize the nuanced details and motion diversity resulting from these properties.

\myparagraph{Intuitive physics guided learning.} The study of intuitive physic~\cite{intuitive1983,WattersZWBPT17,BattagliaPLRK16} has experienced a resurgence with recent attempts to incorporate human-level intuitive physics capabilities into deep learning frameworks. Some work focuses on inferring intrinsic physical properties of objects such as mass or spring~\cite{DuanDFT22,ZhengWPMCH21} while others predict underlying object dynamics such as the trajectory~\cite{MottaghiRGF16,MottaghiBRF16}, forces~\cite{GraviCap2021,wu2017,LererGF16,JannerLFTFW19,GrothFPV18}, or physical stability~\cite{wu2017,LererGF16,JannerLFTFW19,GrothFPV18}. 

Applications of intuitive physics have expanded to 3D computer vision, including recovering object trajectories by modeling gravity ~\cite{GraviCap2021}, and guiding 3D human pose estimation with center of mass and center of pressure~\cite{tripathi2023ipman}.

Our work draws key insights from intuitive physics studies, modeling the intricate interplay between dynamic human forces and object resistance, by explicitly incorporating these physical attributes to generate more realistic and diverse human-object interactions.

\section{Method}
Our goal is to synthesize diverse, nuanced human-object interactions by modeling the physical attributes of resistance and the applied human force. The style of motion can be manipulated not only by varying the resistance, but also by the desired action and contact mode (e.g., left, right, two-handed).

\begin{figure}[t!]
    \centering
    \includegraphics[width=\linewidth]{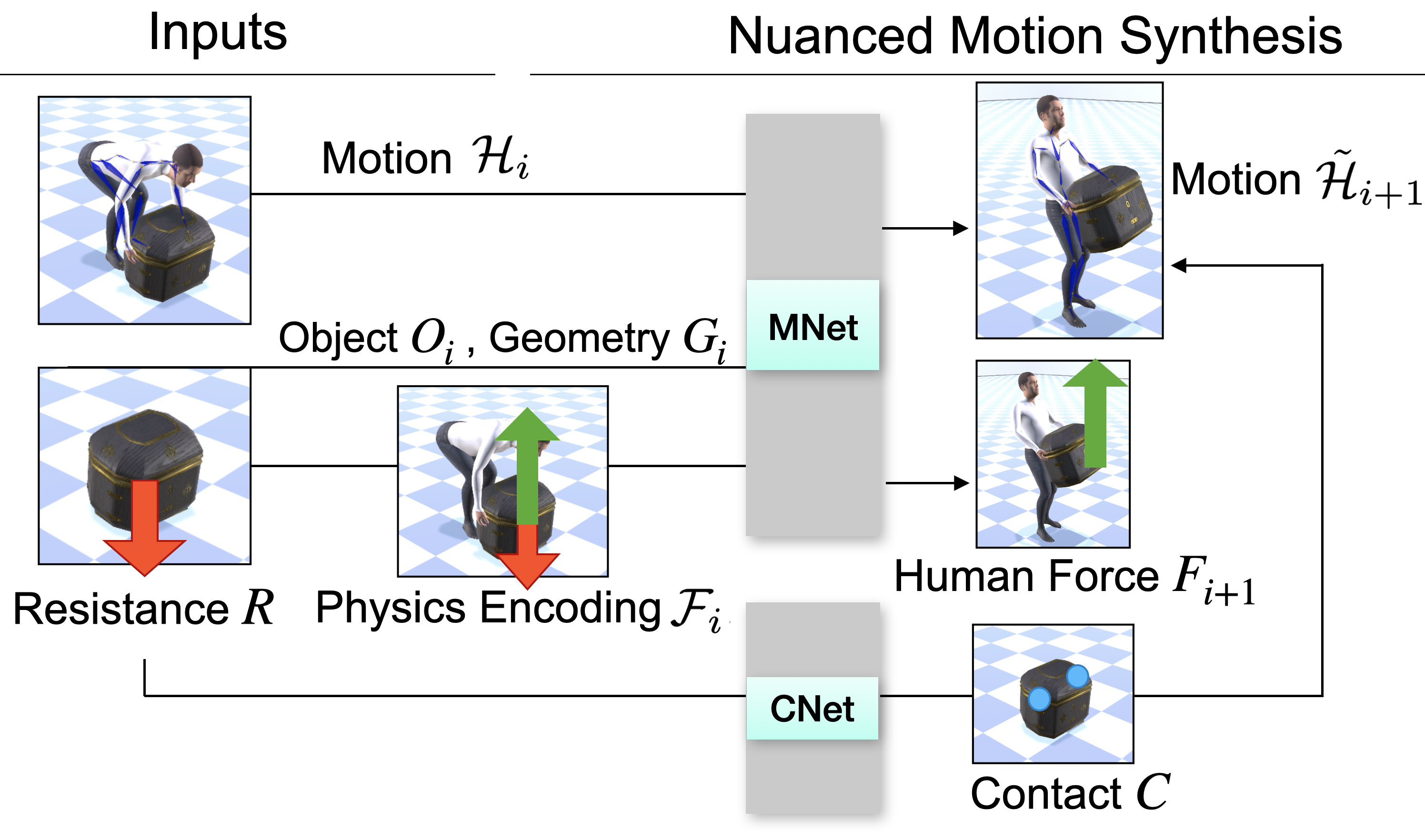}
    \caption{Given the input human and object poses and the object geometry, our method which consists of \emph{MNet} and \emph{CNet}, synthesizes auto-repressively a diverse spectrum of nuanced interactions.}
    \label{fig:architecture}
    \vspace{-15pt}
\end{figure}

\subsection{Key Insight} 
\label{sec:insight}

Learning simultaneously from multi-modal human-object information, including the object pose, different modes of action and contact, can result in motions lacking in diversity and nuanced details. However, the intricate interplay between the force exerted by the human and the perceived resistance can help disambiguate between similar poses and facilities learning from multi-class motions. There are two advantages to utilizing these two physical attributes. a) Human force is a 3D representation that informs the model to which direction the force is applied, which dictates the human motion. b) Given the applied force, the resistance further refines the human motion. For instance, the human leans the body to compensate for the increased resistance.

Founded on this key insight, our method (Figure~\ref{fig:architecture}) utilizes a physics-aware motion network, \emph{MNet} (Section~\ref{sec:mnet}) to auto-regressively synthesize diverse, nuanced interaction motion that is plausible to the physical conditions. More specifically, our method takes as input the initial human pose $\set{H}_0$, the object pose $\goal_0$ and its geometry $\geom_0$, it auto-regressively predicts the future body motion $\set{H}_{i>0}$. Notably, its physics awareness is achieved by our novel physics encoding. To ensure the motion is physically plausible, we use another neural network, \emph{CNet} (Section~\ref{sec:contact}) to predict the hand contact positions on the object surface.

\subsection{MNet: Physics-aware Motion Prediction}
\label{sec:mnet}
At current frame $i$, given the poses of the human $\set{H}_i$ and the object $\goal_i$, the object geometry $\geom_i$, \emph{MNet} leverages the interrelation between the applied human force and the perceived resistance to auto-regressively synthesize the future human motion $\tilde{\set{H}}_{i+1}$. At the core of \emph{MNet}, is the intuitive physics encoding $\set{F}_i$, which encapsulates the crucial physics attributes of the interaction.

 \myparagraph{Intuitive Physics Encoding $\set{F}_i$.} Denoted as $\set{F}_i=\{\force_{i},\resistance, \occupancy_i\}$, the encoding comprises the 3D human force $\force_{i}\in \mathbb{R}^{\tau\times3}$, the magnitude of resistance $\resistance\in \mathbb{R}$, and binary hand contact labels $C\in \{0,1\}^{\tau\times2}$. The control signal is designed according to previous work~\cite{nsm, zhang2022couch,pfnn,mann,starke2021}, the signals are from the $\tau=13$ uniformly sampled frames within the temporal window $[max(0, i-30), i+30]$. This sampling ensures the model's contextual awareness, allowing it to capture the temporal nuances of the interaction. \\
Next, we define the rest of the auto-regressive inputs: \\
\myparagraph{Human motion $\set{H}_i$:} At frame $i$, the human motion comprises of the pose and the root trajectory: $\set{H}_i = \{\pose_{i}, \traj_{i}\}$. The human pose $\pose_{i}= (\poseposition, \posevelocity, \poserotation)$ contains root-relative joint positions $\poseposition\in \mathbb{R} ^{J\times 3}$, velocities $\posevelocity \in \mathbb{R} ^{J \times 3}$, and rotations $\poserotation \in \mathbb{R} ^{J \times 6}$ (forward and upward vectors of the rotation matrix) for the $J=22$ joints in SMPL skeleton~\cite{smpl}.\\
The root trajectory $\mat{T}_{i} = (\trajpos, \trajdir, \trajact)$ encodes the root position $\trajpos\in \mathbb{R} ^{\tau \times 2}$ and direction $\trajdir \in \mathbb{R} ^{\tau \times 2}$ projected onto the ground. $\trajact \in [0, 1] ^{\tau \times 5}$ describes the current action (\emph{idle, walk, carry, push, pull}). We use continuous values between 0 and 1 to facilitate the transition between actions.

\myparagraph{Object $\goal_i$:} The target object is denoted as $\goal_i = (\goalpos, \goaldir, \goalact)$, where $\goalpos \in \mathbb{R} ^{\tau \times 3}$, $\goaldir \in \mathbb{R} ^{\tau \times 6}$ are its positions and orientations relative to the root. $\goalact \in \{0, 1\} ^{\tau \times 5}$ are the binary variables describing the \textit{desired future action}. For example, for carrying, the target action is ``carry'' when approaching the object. 

\myparagraph{Geometry $\geom_i$:} The object geometry $\geom_i$ is encoded by voxelizing object shape (in an $8\times8\times8$ dimensional grid). Each voxel stores its occupancy ($\mathbb{R}$) and the relative vector between the human root joint and the voxel ($\mathbb{R}^3$). This enables reasoning about the shape of the object. We vectorize this grid to obtain our geometry encoding $\geom_i \in \mathbb{R}^{2048}$. 

\myparagraph{Phase $\humanphase_i$:} We introduce a learned variable, phase $\humanphase_{i}\in\mathbb{R}^{4}$, encoding the human joint trajectories following~\cite{deepphase}. It encapsulates the spatial-temporal context of the motion. 

\myparagraph{Training.} 
The \emph{MNet} adopts a mixture-of-experts architecture~\cite{pfnn,samp,nsm,starke2020,zhang2022couch,zhang2023roam}. It is trained with direct supervision by minimizing the MSE loss on the outputs:
\begin{equation}
\begin{split}
\{\tilde{\set{H}}_{i+1}, \force_{i+1}, \goal_{i+1}, \humanphase_{i+1}\} = f^\text{MNet}(\set{H}_i, \set{F}_i, \goal_{i}, \geom_i, \humanphase_i),
 \end{split}
\label{eq:mnet}
\end{equation}
where $\tilde{\set{H}}_{i+1} = \{ \pose_{i+1}, \traj_{i+1}, \futureposeposition, \goaltraj_{i+1} \}$ denotes the future human motion. Here, $\pose_{i+1}$, $\traj_{i+1}$ are the future body joints and the root trajectory. $\force_{i+1}$ denotes the future 3D human forces. $\futureposeposition$ and $\goaltraj_{i+1}$ are joint positions and root trajectory relative to the object. Supervising with these object-centric signals ensures the human reaches the target. $\goal_{i+1}$ and $\humanphase_{i+1}$ are the object pose, and phase at frame $i+1$. More details of the network architecture and training can be found in the supplementary.

\myparagraph{Derivation of Human Force.} 
For training, the human force is not directly observable. However, it can be estimated from the observable object acceleration. a) When the object moves, we leverage Newton's laws to compute the human force. We assume the direction of force acts opposite to the direction of the object acceleration. For example, for \textit{pushing} and \textit{pulling}, the force is derived as $m \cdot g \cdot \eta \cdot \frac{\acceleration}{\|\acceleration\|}+ m \cdot \acceleration$, where $m$, $g$, $\eta$, $\acceleration$ denote the mass, gravity, frictional coefficient and object acceleration. b) When the object is in contact while remaining static, we derive the force by linearly interpolating between the zero, and the derived force from the nearest frame at which the object moves. c) When the resistance is too great for a successful interaction, we leverage the maximal observed human force for each action class as the control signal. Note, to eliminate the effect of noise of object acceleration, we apply a Butterworth filter of kernel size 4.

\begin{figure*}[ht!]
    \centering
    \includegraphics[width=\textwidth]{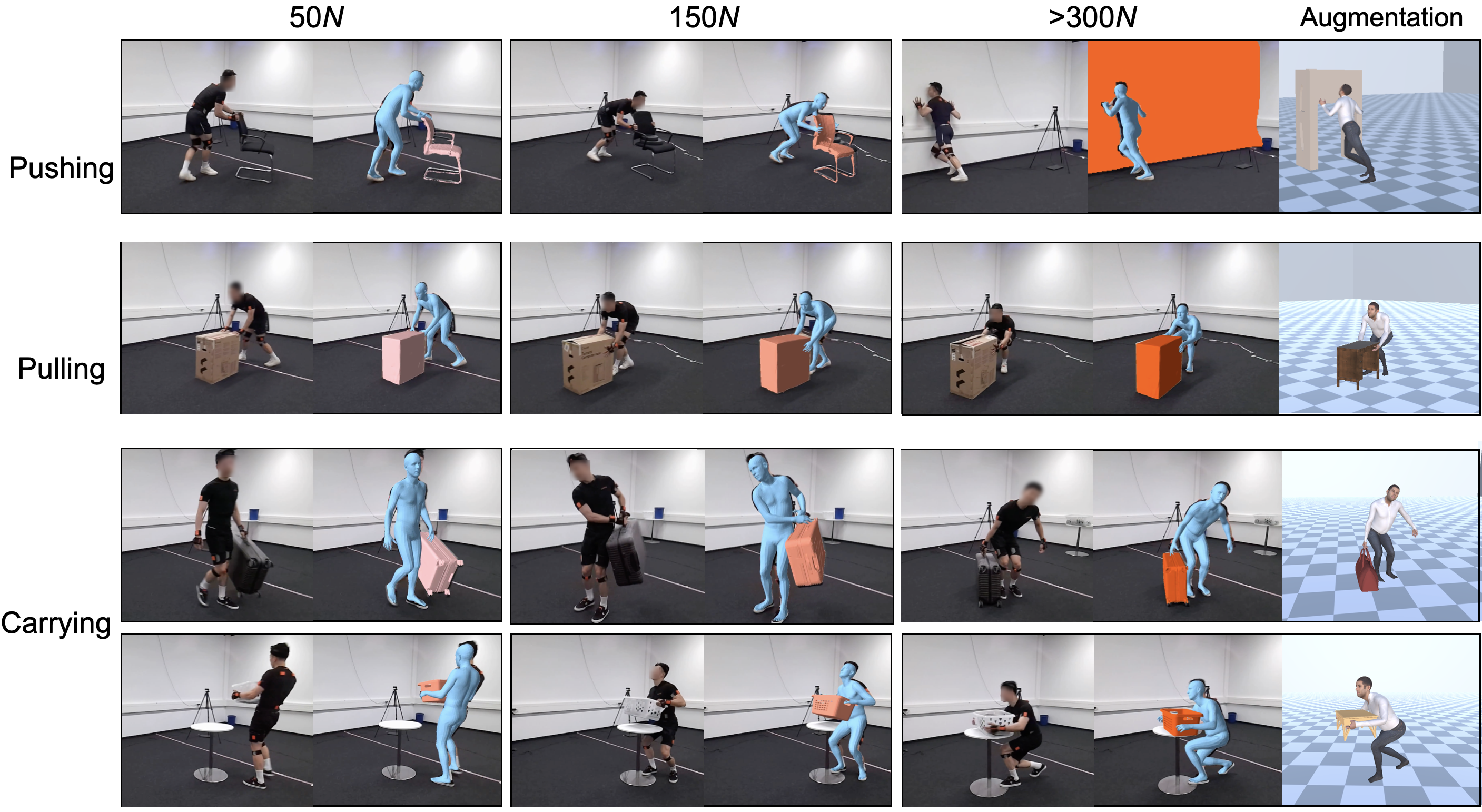}
    \caption{The FORCE dataset accurately captures diverse, nuanced interaction motion under varying levels of resistance ($N$ for Newtons). The last column shows results from the object shape augmentation (see Section~\ref{sec:dataset}), where nuanced motion details are preserved.}
    \label{fig:data}
    \vspace{-8pt}
\end{figure*}

\subsection{Resistance-conditioned Contact Prediction}
\label{sec:contact}
Contacts are imperative to ensure the physical plausibility of the human-object interaction, and the way the human contacts with an object is influenced by its resistance. For example, when pushing a chair when the resistance is low, the human tends to push through the top of the chair since it is more efficient. However, when the resistance of the chair increases, the human tends to hold onto the lower part of the chair to stabilize to push more efficiently. Inspired by this and the previous work on predicting hand contact conditioned on the object geometry~\cite{zhang2022couch}, we use a variational auto-encoder \emph{CNet} to predict the hand contact positions, $\set{C}\in \mathbb{R} ^{2\times3}$, on the object surface. Formally, the contacts are predicted via $\hat{\set{C}} = f^\text{CNet}(\resistance, \geom_i, \poseposition, \goalact)$. Here, $\resistance$,  $\poseposition$, $\goalact$ are the resistive force, human joint positions and the target actions as defined previously. 
The object geometry $\geom_i$ in this case is encoded relative to the center of the object instead of the human's root. During training, the network is trained to minimize the following loss, 
\begin{equation}
\begin{split}
 L_{\mathrm{CNet}} = \|\hat{\set{C}} - \set{C}\|_{2}^{2}  + \lambda KL(q(\latent | \contact, \resistance, \geom, \poseposition, \goalact )\| p(\latent)),
  \end{split}
\label{eq:contact}
\end{equation}
where $KL$ denotes the Kullback-Leibler divergence. For training, \emph{CNet} is trained with ground truth contact positions derived from Section~\ref{sec:dataset}. For ``one-handed'' contact, the contact position for the other hand is zeroed. During inference, the contacts are sampled once the human is within the vicinity of the object, and is remained constant for the rest of the motion sequence. We apply a real-time inverse kinematics~\cite{samp,nsm,zhang2022couch,zhang2023roam} to ensure the contact constraints are satisfied. We optimize the pose of the arm joints to minimize the distance between the linearly interpolated position between the current hand joint and contact. The weight of the interpolation is inversely proportional to the distance to ensure the smoothness and realism of the hand motion.

\section{Dataset}
\label{sec:dataset} 

% Our goal is to build a model of human-object interactions that is aware of the intuitive physics, yet existing datasets fall short in accurately capturing human interactions with moving objects~\cite{nsm,samp,zhang2022couch,bhatnagar22behave,huang2022intercap}. A recent dataset offers high-quality dynamic interaction data, but it does not capture different styled human motion under varying physical conditions~\cite{li2023OMOMO} .
% To address this gap, w
We introduce a dataset that comprises $450$ motion sequences of human pushing, pulling and carrying.  Its main characteristic is the diverse motion nuances through interactions with varying resistances. It includes 8 objects, captured with 3-6 levels of resistance (from 5kg to 35kg) with removable weights. The testing set consists of 30 sequences covering a range of interactions and scenarios.

% \begin{table}[t!]
%     \small
%     	\caption{Comparison with existing human-object interaction datasets.} 
% 	\centering
% 	\resizebox{\columnwidth}{!}{
% 	\begin{tabular}[b]{c|c|c|c|c}
% 		\hline
% 		{Dataset} & {Goal-reaching} & {Dynamic Obj.} & {RGB-D}   & {Resistance}   \\
%         \hline
%         $\textrm{NSM~\cite{nsm}}, \textrm{SAMP~\cite{samp}}$ & \greencheck & \redcross & \redcross & \redcross \\
%         $\textrm{COUCH~\cite{zhang2022couch}}$ & \greencheck & \redcross & \greencheck & \redcross \\
%         $\textrm{BEHAVE~\cite{bhatnagar22behave}}, \textrm{InterCap~\cite{huang2022intercap}}$ & \redcross & \greencheck & \greencheck& \redcross \\
%         $\textrm{OMOMO~\cite{li2023OMOMO}}$ & \redcross & \greencheck & \redcross & \redcross \\

%       	\hline

%           $\textrm{Ours}$ & \greencheck & \greencheck & \greencheck & \greencheck\\

%     	\hline
% 	\end{tabular}
% 	}
%     \label{table:dataset}
%     \vspace{-17pt}
% \end{table}

 \myparagraph{Human and object tracking.} We leverage a customized tracker, integrating 17 human-mounted IMU~\cite{xsens} sensors with the cameras. This enhancement improves the accuracy of the captured data. After initially fitting the SMPL\cite{smpl} to the captured point clouds, we synchronize the tracking results with motion sequences recorded by IMU sensors. Subsequently, we optimize the body poses from the IMU tracking to align with the Kinect-fitted result. \\
 For objects, we initiate the tracking by pre-scanning eight objects (carry: bin, suitcase, stool, basket, container, backpack; push/pull: chair, box) to obtain their template meshes. Subsequently, we fit them to annotated keypoints on the captured images using camera projection. The fitting is refined by running ICP to fit to the segmented object point cloud~\cite{cheng2022xmem}. We consider our objects to be rigid.\\
 Further technical details about our tracking methodology are available in the supplementary materials.

\myparagraph{Diverse interactions with varying resistance.} The capture covers diverse motion variation. Firstly, we vary the resistance of the object in each interaction instance. The same motion is performed for each object of the same action category. Secondly, the dataset captures different contact modes including one-handed and two-handed interactions. We systematically place them in varying positions relative to the human. To ensure the authenticity of the motion, only high-level guidance is provided instead of specific instructions. Figure~\ref{fig:data} showcases a selection of interactions.  \\
Note, that our data collection is performed on a uniform surface, and we determine the frictional coefficient by measuring force meter readings during the uniform pulling of a 10kg object. For consistency, we approximate gravity ($g$) and the frictional coefficient ($\eta$) as 10N/kg and 1, respectively.

\begin{figure*}[h!]
    \centering
    \includegraphics[width=0.8\textwidth]{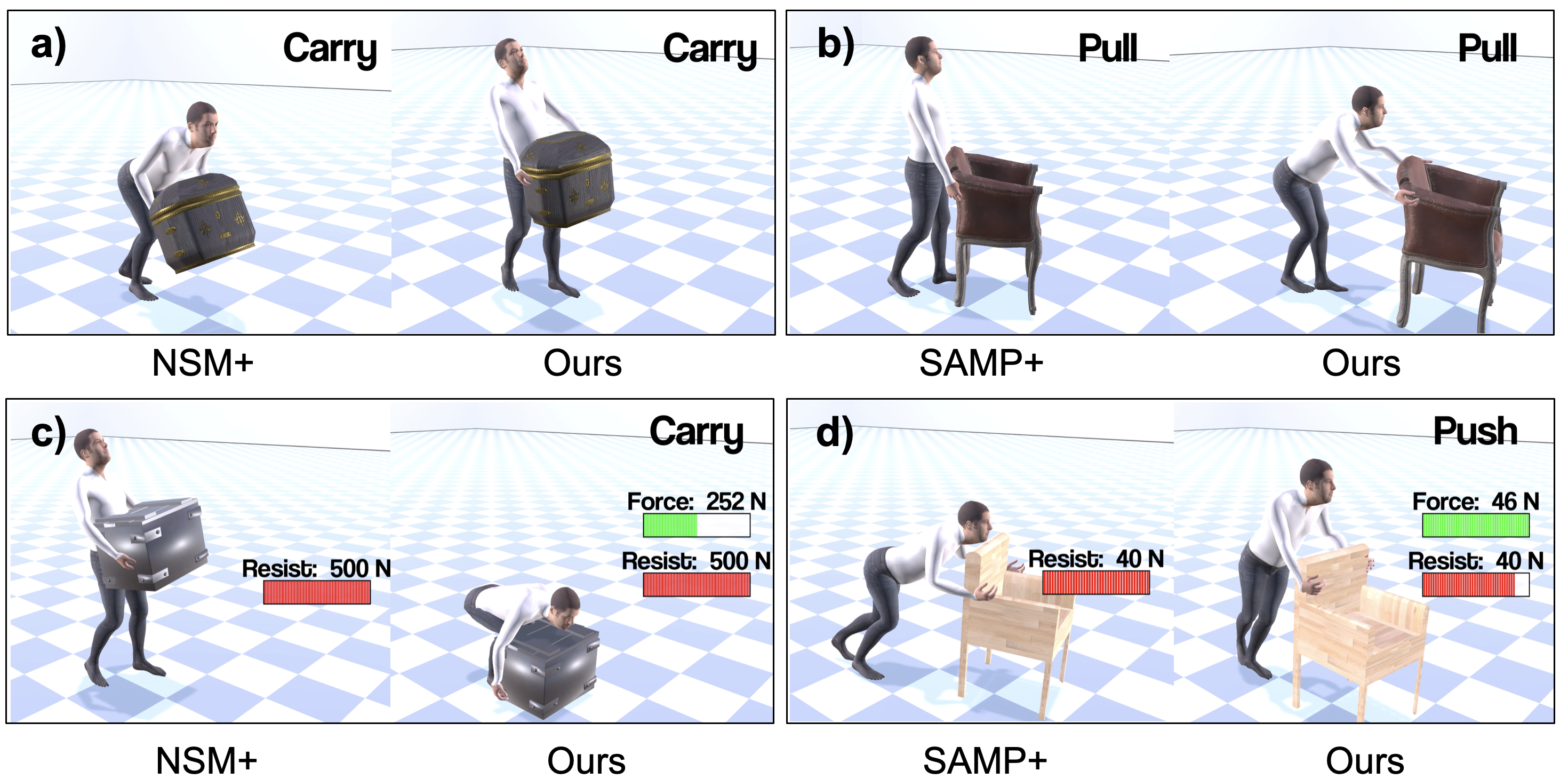}
    \caption{Comparison with baselines. Our \emph{intuitive physics encoding} $\set{F}$ disambiguates different multi-class motions and enables the synthesis of the desired interaction (carrying in \textbf{a}, pulling in \textbf{b}). \textbf{c)} and \textbf{d)} show the coupled encoding of human force and resistance allows the synthesis of visually plausible interactions. In \textbf{c}), when the resistance is greater than the human force, \model synthesizes a more plausible ``infeasible'' interaction. In \textbf{d)}, with low resistance, the human pose maintains an upright position.}
    \label{fig:baseline}
    \vspace{-10pt}
\end{figure*}

\myparagraph{Object shape augmentation.} To generalize to unseen shapes, each motion sequence is augmented with 10 differently shaped objects from one of the table, chair, bag, and shelf categories of ShapeNet~\cite{shapenet} (see the last column of Figure~\ref{fig:data}). First, we detect the ground truth hand contact positions on the object surface based on a distance threshold of 0.05cm. Next, we transfer contacts by aligning the new object with the source via centering and scaling. The ground truth contacts are projected onto the nearest surface of the new object. At every frame, following object sampling and rescaling, the human poses are recomputed to satisfy the contacts using a full-body inverse kinematics~\cite{nsm}. We assume the resistance remains unchanged after augmentation to preserve the context of human motion.

% \myparagraph{How does this dataset benefit the community?}
% \begin{itemize}
% \setlength\itemsep{0.1cm}
%     % \item Training and evaluating physics-based synthesis methods~\cite{hassan2023,xiao2023unified,pan2023interscene}.
%     \item Developing physics-based synthesis methods~\cite{hassan2023,xiao2023unified,pan2023interscene}.
%     \item Synthesizing non-physics-informed object interactions~\cite{nsm,samp,li2023OMOMO,zhang2022couch,xu2023interdiff}.
%     \item Human-object reconstruction~\cite{xie22chore, xie2023vistracker, petrov2023popup,GraviCap2021}.
%     \item Benchmarking pose estimation for occlusion scenarios~\cite{ChengYWT20,sarandi2018,ChengYWWT19}.
%     \item Tracking human-object interaction with RGB-D~\cite{bhatnagar22behave,Huang:CVPR:2022}.

% \end{itemize}

\section{Experiments}
\label{sec:experiments}

\begin{figure*}[t!]
    \centering
    \includegraphics[width=.8\textwidth]{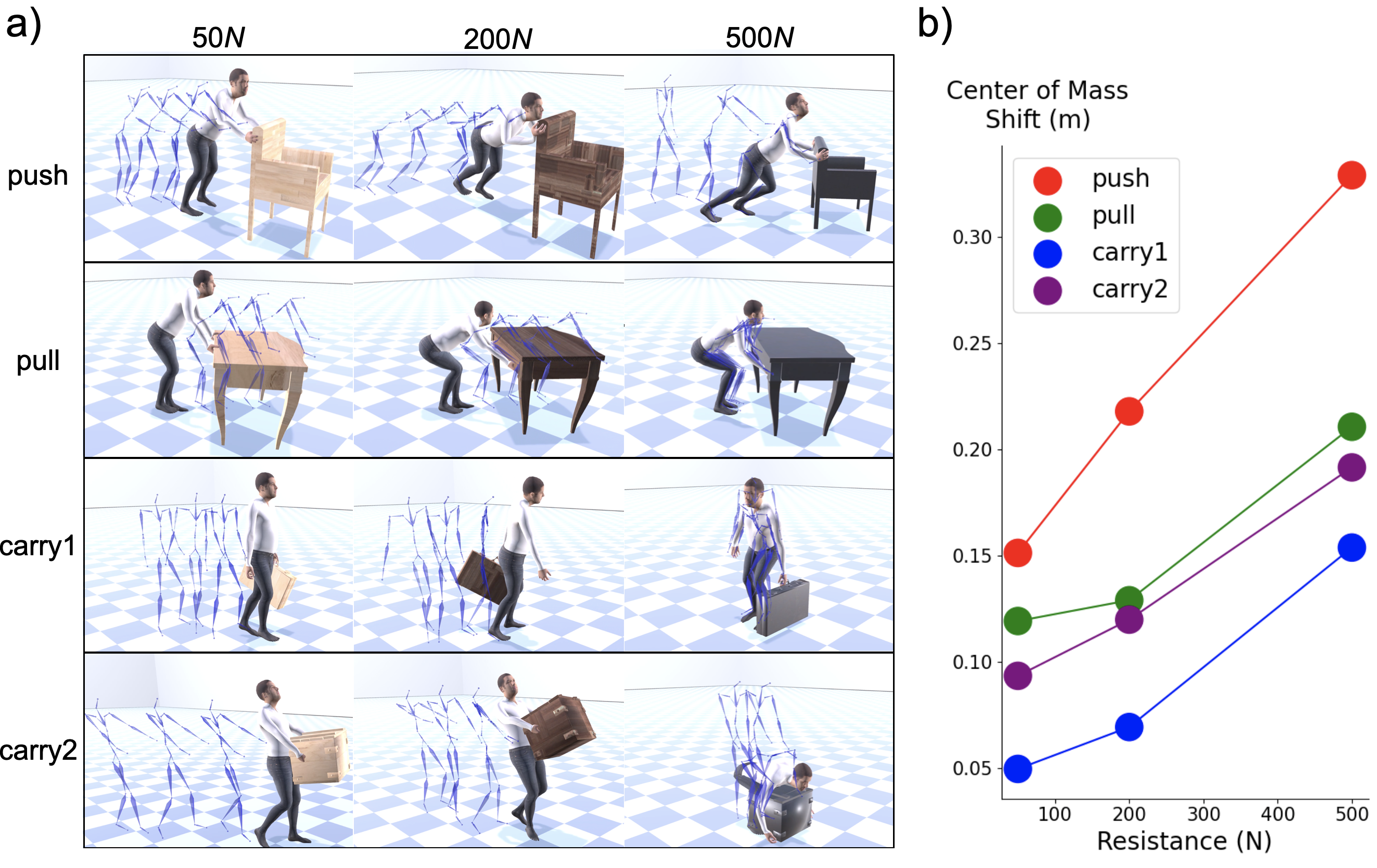}
    
    \caption{Nuanced interactions with the same object shape, but different resistance. a) The human struggles with the interaction as the resistance increases from 50~\textit{N} to 200~\textit{N}. At 500~\textit{N}, the human fails to interact with the object. b) Plot depicts the horizontal shift of the center of mass of the human, estimated following~\cite{tripathi2023ipman,com}.  Results averaged over 1000 frames of the synthesized interaction.}
    \label{fig:resistance}
    % \vspace{-10pt}
\end{figure*}

For evaluation, we first introduce the baselines (Section \ref{subsec:baselines}). We then compare with the baselines quantitatively (Section \ref{subsec:quant}) and qualitatively (Section \ref{subsec:quali}). Last, we perform ablations on the designs of our method  (Section \ref{subsec:ablation}) and show results on generalization (Section \ref{subsec:generalisation}). \textit{We encourage readers to refer to our supplementary video for animated qualitative results.}
\vspace{-5pt}

\subsection{Baselines}
\label{subsec:baselines}

To our knowledge, the most related works to ours are NSM \cite{nsm} and SAMP \cite{samp} as they both enable real-time generation human-object interaction with a user-controller. We also compare our method with InterDiff~\cite{xu2023interdiff}, a state-of-the-art method that leverages the diffusion model. 

We train the baseline methods on our captured dataset using their publicly released codes. These methods neglect the physics attributes of the interaction. To measure the effectiveness of our intuitive physics encoding, we adapt NSM and SAMP by additionally conditioning on the resistive force and the binary hand contact. We refer to these modified baselines as NSM+ and SAMP+. Moreover, we evaluate two ablative variations of~\model\removespace. One model is without the physics encoding $\set{F}$, and the other one is without the contact prediction. 
\vspace{-5pt}

\subsection{Quantitative Evaluation} 
\label{subsec:quant}

\begin{table}[ht!]
    \small
     	\caption{Quantitative comparisons for offline motion synthesis. Results evaluated on the testing sequences.}
	\centering
	\resizebox{\columnwidth}{!}{
   \begin{tabular}{lccc}
        \hline
        Method & $\text{MPJPE(cm)}^{\downarrow}$ & $\text{Non-collision($\%$)}^{\uparrow}$ & $\text{Foot-slide(cm)}^{\downarrow}$ \\
        \hline
        NSM \cite{nsm}   &  7.71  & 64.3 &  4.59 \\
        SAMP \cite{samp} &  8.67  & 57.7 &  5.79 \\
        InterDiff~\cite{xu2023interdiff} & 8.62 & 75.3 & 6.82 \\
        \hline
        NSM+   &  7.08  & 69.6 &  4.71  \\
        SAMP+  &  8.34  & 58.3 &  6.02 \\
        \hline
        Ours no Phys. Enc. $\set{F}$  & 6.75 & 75.4 &  4.60 \\
        Ours no Contact Pred.    & -  & 77.2 &  4.48 \\
        Ours no \textbf{R} & 7.10 & 76.4 & 4.76 \\
        Ours             & \textbf{6.02} & \textbf{84.0} &  \textbf{4.46} \\
        \hline
    \end{tabular}
	}

    \label{table:quant1}
    \vspace{-5pt}
\end{table}

\begin{table}[ht!]
    \small
     	\caption{Quantitative comparisons for online full interaction synthesis. Note, diversity on the dataset is \textit{0.906}. The closeness to this figure indicates a better result.}
	\centering
	\resizebox{\columnwidth}{!}{

    \begin{tabular}{lccc}
        \hline
        Method  & $\text{Diversity}^{\rightarrow}$ & $\text{Success Rate($\%$)}^{\uparrow}$ & $\text{Exec. Time(s)}^{\downarrow}$ \\
        \hline
        NSM \cite{nsm}   & 0.707&  75.1   & 5.94 \\
        SAMP \cite{samp} & 0.719&  66.2   & 7.72 \\
        \hline
        NSM+   & 0.785&  79.6   & 6.01 \\
        SAMP+  & 0.776&  68.7   & 7.56 \\
        \hline
        Ours no Phys. Enc. $\set{F}$  & 0.847 & 86.7  & 5.81 \\
        Ours no Contact Pred.    & 0.886 &  94.2  & 5.50 \\
        Ours           & \textbf{0.891}  & \textbf{97.5}  & \textbf{5.42} \\
        \hline
    \end{tabular}
   
	}

    \label{table:quant2}
    \vspace{-15pt}
\end{table} 

We quantitatively evaluate our model performance with two setups. \textbf{a)} offline motion synthesis. This is performed on the testing sequences given the ground truth control signals. We evaluate the accuracy (MPJPE) and motion quality (foot-slide and collision) of motion synthesis on testing sequences following our training/testing split explained in Section~\ref{sec:dataset}. The model synthesizes poses auto-regressively given the trajectories. The sequences are each distinctive, covering diverse pushing, pulling and carrying motions at different resistances. \textbf{b)} online full synthesis. We synthesize full approaching and interacting in a random setup. For fairness, 10 objects are randomly sampled. They are placed randomly relative to the human (within 6 meters) at a random orientation. The synthesis is performed 120 times for each method, repeating each action type (pushing, pulling, one-handed, and two-handed carrying) 30 times. We measure the diversity, the success rate and the execution time. Note, InterDiff~\cite{xu2023interdiff} does not allow online synthesis, we compare against it for offline synthesis on the testing set.

\myparagraph{MPJPE(cm)\rm{~\cite{quaternet,h36mpami}}.}  The accuracy of human motion synthesis. The result on ``Ours no Contact'' is omitted because the ablative baseline utilizes the same MNet as ``Ours''. \myparagraph{Non-Collision Score($\%$)\rm{~\cite{Zhang2020CVPR, PLACE:3DV:2020, Hassan:CVPR:2021}}.} The percentage of frames that the body does not penetrate with the object. \\ \myparagraph{Foot-slide(cm)\rm{~\cite{pfnn, nsm, hassan2023, samp, zhang2023roam, zhang2022couch}}.} The average traveled distance by the pivotal foot per step is calculated.  \\ \myparagraph{Diversity\rm{~\cite{Guo2022CVPR,yuan2020dlow,Zhang2020CVPR,samp}}.} We compute the Average Pairwise Distance (APD) on the root-relative joint positions. \\\myparagraph{Success rate($\%$)\rm{~\cite{learningtosit,samp,hassan2023}}.} 
The percentage of both the desired object interaction and the hand contact is synthesized within 10 seconds.\\ \myparagraph{Execution time\rm{~\cite{Hassan:CVPR:2021,pan2023interscene}}.} Time till a successful interaction. 

From Table~\ref{table:quant1}, it can be seen that ~\model attains the lowest MPJPE of $6.02$ cm, surpassing both baseline models and ablative variations of \model. This emphasizes the crucial role of our intuitive physics encoding in facilitating the learning of multi-class motions. Additionally, our method minimizes object interpenetration, as reflected in the highest non-collision scores. It is notable that, compared with the ablative baseline without contact prediction, our method has a higher non-collision score. This indicates that with contact prediction, our model synthesizes more physically plausible motion. Furthermore, our model demonstrates improved motion quality, evident in its reduced foot-slide.\\
From Table~\ref{table:quant2}, it can be seen under the online synthesis setup,~\model has the highest success rate of interaction of $97.5\%$ and the shortest average execution time of $5.42$ seconds, illustrating the better controllability of our approach. Our model synthesizes motion of diversity (0.891) which is the closest match to the training data distribution (0.906). This underscores the significance of the novel physics encoding in enabling the synthesis of diverse, nuanced interactions.

\subsection{Qualitative Evaluation} 
\label{subsec:quali}
Figure~\ref{fig:baseline} illustrates our qualitative comparison with the baselines NSM+~\cite{nsm} and SAMP+~\cite{samp}. In Figure~\ref{fig:baseline} a), the poses when lifting up an object and pushing an object are similar, introducing ambiguity between these two classes of motions. However, the physics encoding informs the model about the crucial directional information of the human force, dictating the model to synthesize the desired carrying motion. Similarly in Figure~\ref{fig:baseline} b), our model synthesizes the desired pulling motion. The physics encoding also provides awareness of the feasibility given the resistance, because of the coupled encoding of human force and resistances. In Figure~\ref{fig:baseline} c), when the resistance is greater than the exerted human force, \model is aware that the interaction is no longer feasible. In Figure~\ref{fig:baseline} d), our synthesis is more visually plausible given the low resistance, with the body pose remaining upright.

\begin{figure}[h!]
    \centering
    \includegraphics[width=.8\linewidth]{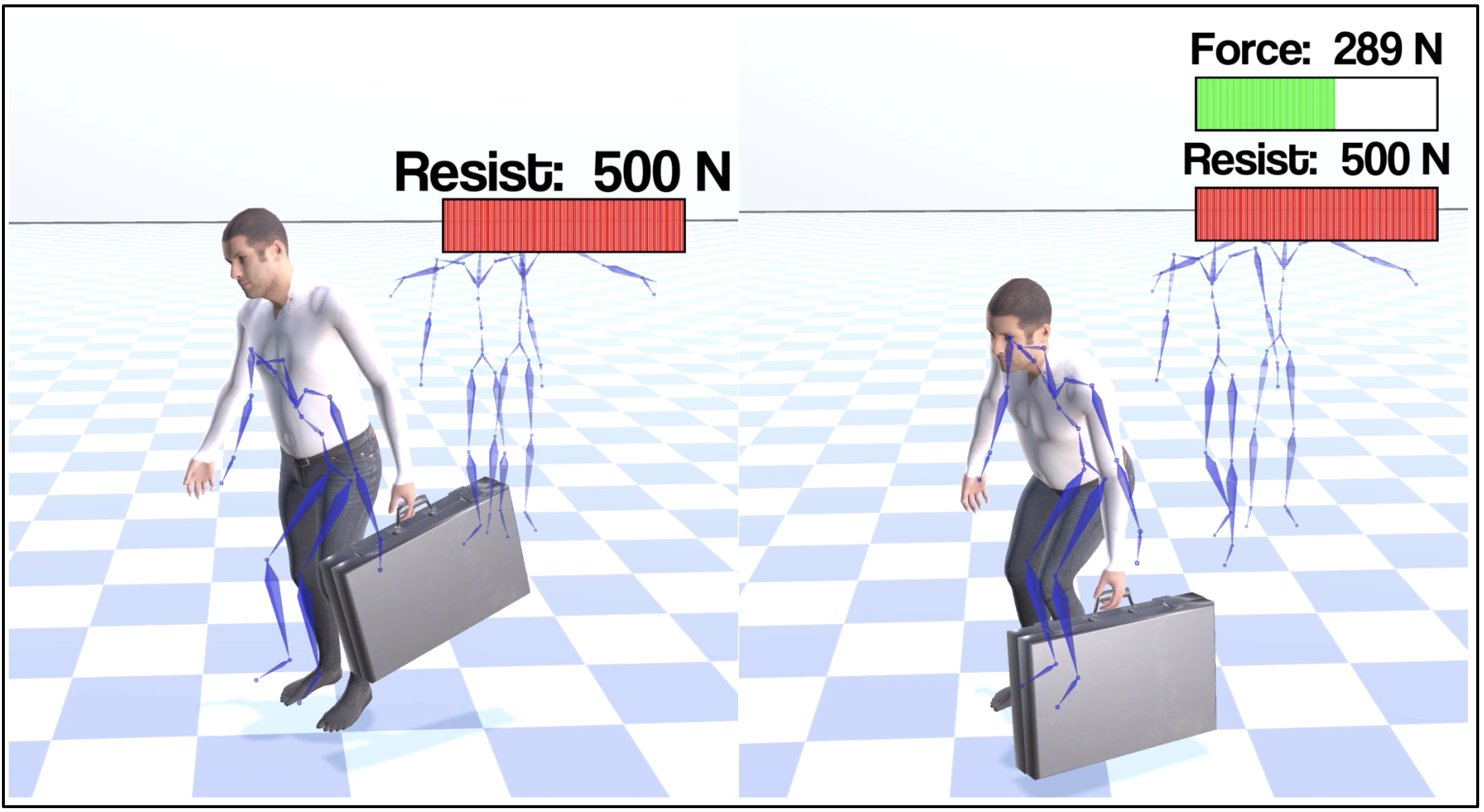}
    \caption{Without Physics Encoding (left) vs. FORCE (right). With physics encoding, synthesizes visually plausible motion, when the resistance is greater than the applied human force. }
    \label{fig:phys}
    \vspace{-10pt}
\end{figure}

\begin{figure}[h!]
    \centering
    \includegraphics[width=.8\linewidth]{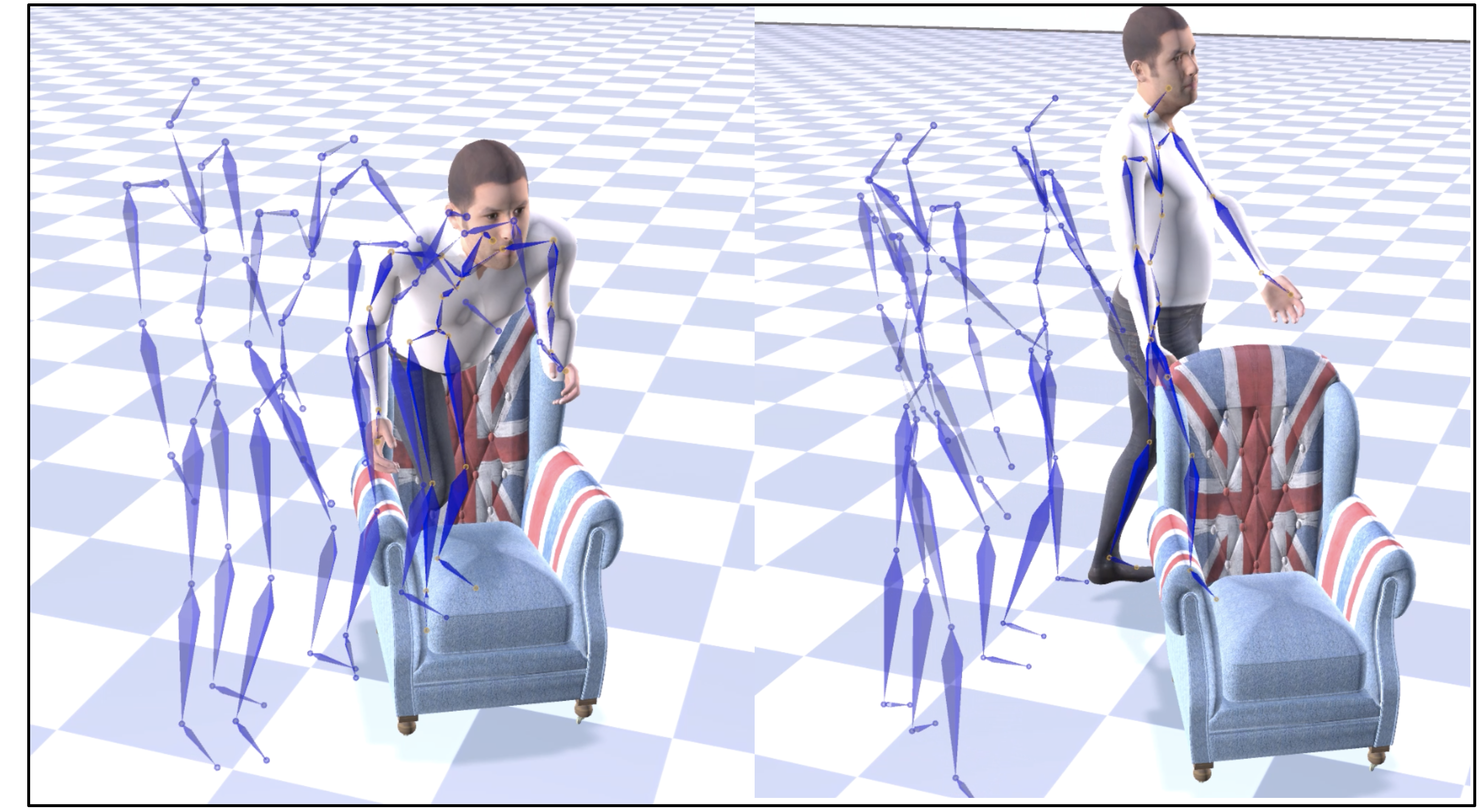}
    \caption{Without geometry representation $\geom$ (left) vs. FORCE (right). The feature enables the reasoning about collision and avoid interpenetration.}
    \label{fig:scene}
    \vspace{-10pt}
\end{figure}

% \begin{figure}[h]
%     \begin{minipage}[c]{0.50\linewidth}
%     \includegraphics[width=\linewidth]{Figures/Figure10.png}
%     \caption{Without Physics Encoding (left) vs. FORCE (right). With physics encoding, synthesizes visually plausible motion, when the resistance is greater than the applied human force. }
%     \label{fig:phys}
%     \vspace{-10pt}
%     \end{minipage}
% \hfill
%     \begin{minipage}[c]{0.48\linewidth}
%     \includegraphics[width=\linewidth]{Figures/Figure10.png}
%     \caption{Without geometry (left) vs. FORCE (right). With geometry representation $\geom$, reasons about collision and to avoid interpenetration.}
%     \label{fig:scene}
%     \end{minipage}%
%    \vspace{-10pt}
% \end{figure}

\subsection{Ablations}
\label{subsec:ablation}

\myparagraph{Physics encoding $\set{F}$.} As shown in Figure~\ref{fig:resistance},~\model synthesizes nuanced human-object interactions with objects of the same shape but varying resistances. Empowered by the intuitive physics encoding $\set{F}$, our method synthesizes diverse interactions. It generalizes not only to different resistances of the object but also to different actions and contacts. The graph on the right provides a zoomed-in analysis of the motion, where we evaluate the horizontal shift of the center of mass of the human (for each 1000-frame sequence). The center of mass of the human is estimated following~\cite{tripathi2023ipman,com}. It can be observed from the plot, that there is a direct positive correlation between the perceived resistance of the human during the interaction and the shift of center of mass. 

Figure~\ref{fig:phys} also emphasizes the importance of intuitive physics encoding. On the left, without the physics encoding, the model overlooks the resistance surpassing the applied force of the human. With the physics encoding on the right, our method is aware that the resistance is not feasible, as the perceived resistance reaches 500~\textit{N} and is greater than the applied human force.

\myparagraph{Resistance $\set{R}$.} As shown in table~\ref{table:quant1}, without conditioning on the resistance, the motion quality degrades.

\myparagraph{Geometry representation $\geom$.} With the representation introduced in Section~\ref{sec:mnet}, our method reasons about the collision as the voxels provide coarse geometry details. As a result, it exhibits less interpenetration (Figure~\ref{fig:scene}).

\begin{figure}[ht!]
    \centering
    \includegraphics[width=\linewidth]{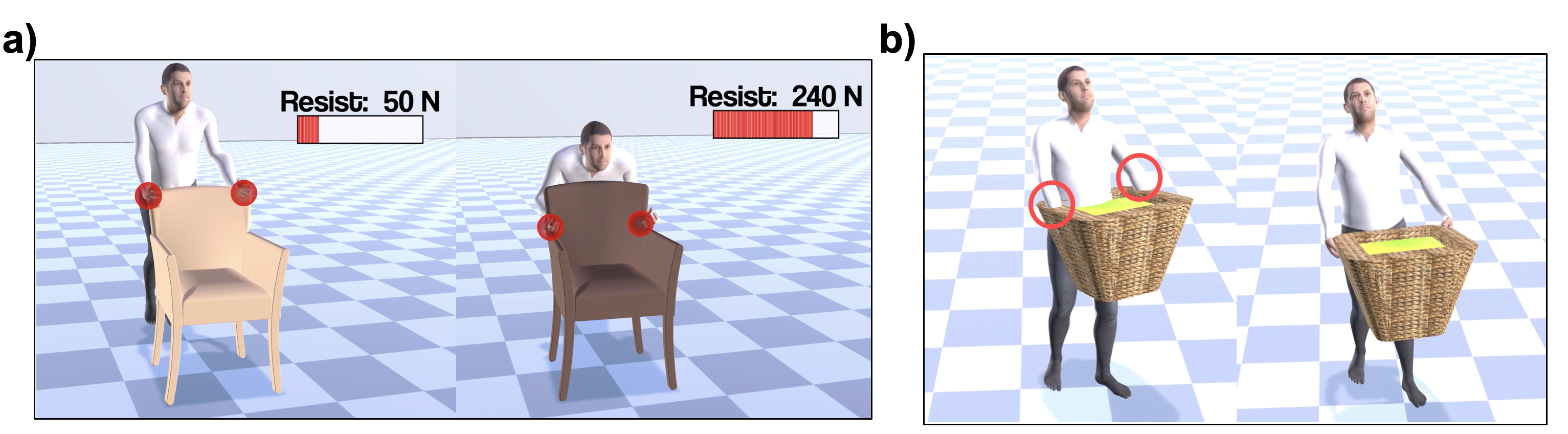}
    \caption{a) FORCE predicts different hand contacts at different resistances for the same pushing motion. b) No contact prediction (left) vs. FORCE (right). FORCE exhibits less penetration.}
    \label{fig:contact}
    % \vspace{-15pt}
\end{figure}

\myparagraph{Contact prediction.} As shown Figure~\ref{fig:contact} a), different contact positions are predicted by conditioning on different resistances using our method. It can be concluded that the contact depends on the resistance. Here, as the resistance increases, the human tends to push from the lower part of the chair to push more efficiently. Figure~\ref{fig:contact} b) shows, with the contact prediction, \model synthesizes motion that satisfies better the hand contact constraints on the surface, and exhibits less interpenetration (see Table~\ref{table:quant1}).

\subsection{Generalization to shapes and locations.}
\label{subsec:generalisation}
\begin{figure}[h!]
    \centering
    \includegraphics[width=\linewidth]{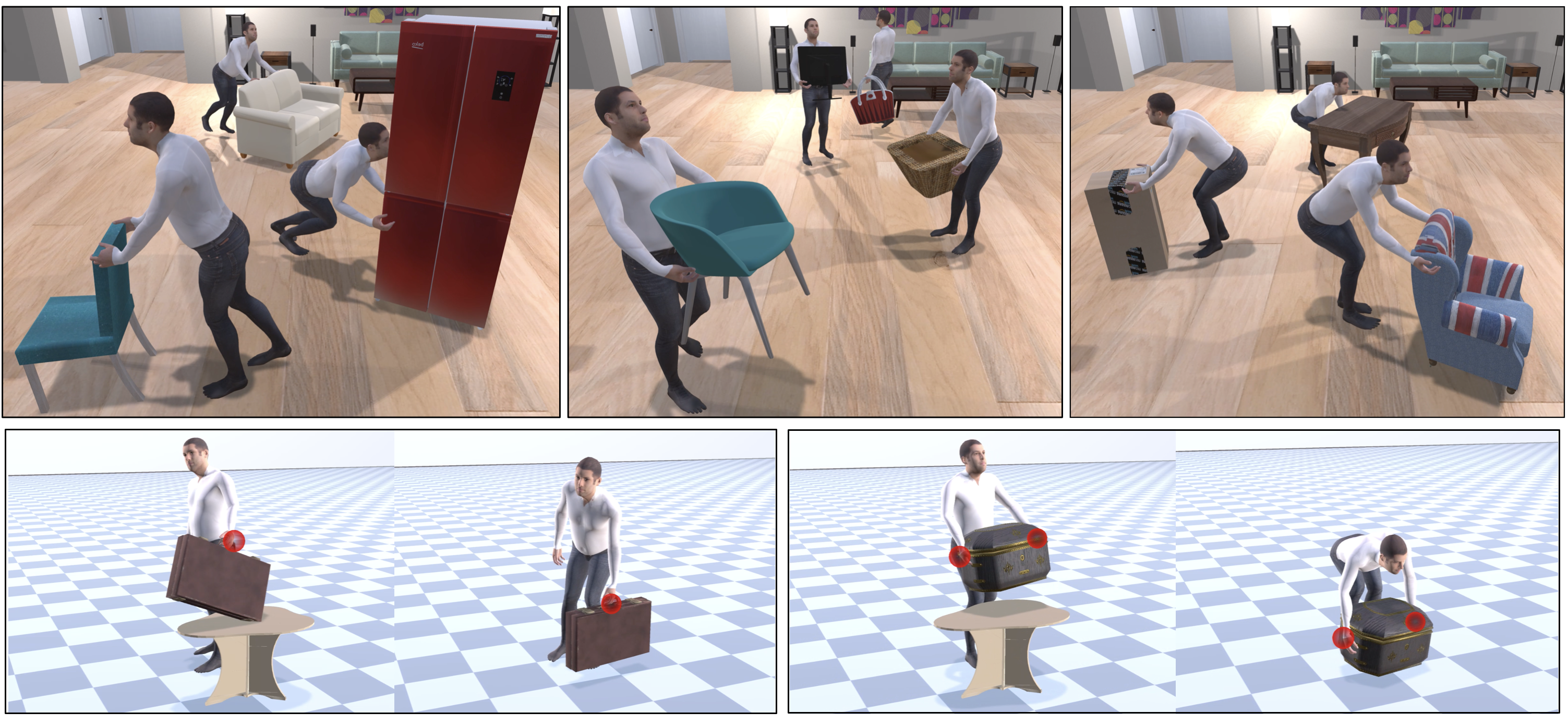}
    \caption{(Top) \model generalizes to unseen shapes, demonstrated via pushing, carrying, and pulling. (Bottom) \model generalizes to different locations, performing one-handed and two-handed carry from the table and from the floor.}
    \label{fig:generalisation1}
    % \vspace{-15pt}
\end{figure}

As illustrated in Figure~\ref{fig:generalisation1},~\model generalizes to different shapes. Thanks to the object shape augmentation introduced in Section~\ref{sec:dataset}, our model is able to synthesize lifting a chair, an object category that is not seen during training of carrying sequences. It is also highlighted in Figure~\ref{fig:generalisation1} that~\model interacts with objects at different locations, for example, picking up from different heights. 

% \begin{figure}[h!]
%     \centering
%     \includegraphics[width=\linewidth]{Figures/Figure7.png}
%     \caption{}
%     \label{fig:generalisation2}
%     \vspace{-8pt}
% \end{figure}

% \input{FORCE-arxiv/6_conclusion}
\section{Conclusion}
This work, \model\removespace, tackles the problem of synthesizing human object interactions with nuanced details, by modeling intuitive physics. It solves two main challenges. First, it proposes the first kinematic-based method that synthesizes human-object interaction conditioned on physical attributes such as object resistance and human force. By leveraging the novel physics encoding, our method generates a diverse spectrum of interaction motions under varying resistance. Its performance in diversity and realism surpasses previous methods. Our accompanying dataset features over 450 motion sequences. Together, they provide a benchmark for training and evaluating HOI methods.

For limitations and future work, this work paved the way for research in nuanced human-object interaction in more complex scenarios. First, while our model demonstrates its efficacy in generating diverse human-object interactions, there is an opportunity for expansion by considering a broader range of subjects, because each individual possesses unique strengths and approaches to interactions. Second, interaction can be extended to complex scenarios of dynamic resistance. For example, when carrying a tank of water, the resistance fluctuates and influences human motion. We release the code, dataset, and models, to stimulate future research.

\section*{Acknowledgement}
We thank RVH group members for their helpful discussions. This work is funded by the Deutsche Forschungsgemeinschaft (DFG, German Research Foundation) - 409792180 (Emmy Noether Programme, project: Real Virtual Humans), and German Federal Ministry of Education and Research (BMBF): T{\"u}bingen AI Center, FKZ: 01IS18039A, and Huawei Noah's Ark Lab. Gerard Pons-Moll is a member of the Machine Learning Cluster of Excellence, EXC number 2064/1 – Project number 390727645. The project was made possible by funding from the Carl Zeiss Foundation.
\newpage

\bibliographystyle{splncs04}
\bibliography{Main}
\newpage
\section*{APPENDIX}

% In this document, we provide explanations for the list of symbols (Section~\ref{sec:symbol}). Subsequently, we offer additional details on the \model dataset (Section~\ref{sec:data}), as well as insights into the architectures and training process (Section~\ref{sec:training}). Finally, we delve into the ethical considerations, potential biases, and broader impact of the work (Section~\ref{sec:bias}).

% \bibliographystyle{splncs04}
% \bibliography{Main}
% \newpage
% \section*{APPENDIX}

% In this document, we provide explanations for the list of symbols (Section~\ref{sec:symbol}). Subsequently, we offer additional details on the \model dataset (Section~\ref{sec:data}), as well as insights into the architectures and training process (Section~\ref{sec:training}). Finally, we delve into the ethical considerations, potential biases, and broader impact of the work (Section~\ref{sec:bias}).

\section*{1\quad Failure Case}
\label{sec:f}
\begin{figure}[h]
    \centering
\includegraphics[width=0.5\linewidth]{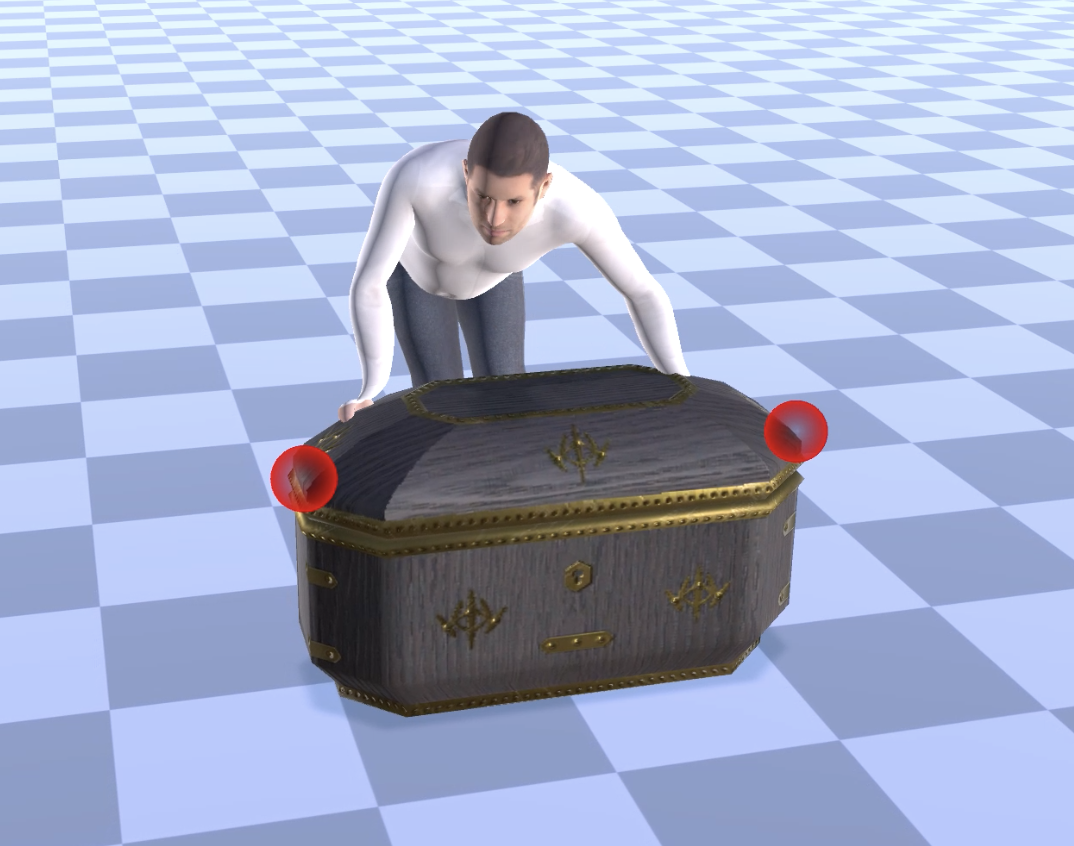}
\caption{The interaction exhibits artifact when the object shape is too large.}
\label{figure:f}
\end{figure}
The object augmentation that we employed for training enables FORCE to generalize to unseen object shapes at testing (see all the animations in the supplementary video.) However, when the object shape is too large, there may exist interpenetration artifact. 
% \end{figure}

\section*{2\quad Dataset}
\label{sec:data}
\myparagraph{Details on the dataset:} Our dataset comprises 450 motion sequences involving human-object interactions with a diverse range of resistance forces. Table~\ref{table:weight} provides a detailed breakdown of the dataset categorized by the level of resistance. In this context, resistance is measured solely by the mass of the removable weight used. The masses of the objects themselves are not factored into these measurements. They are measured and will be provided with the dataset. The dataset distributions based on the type of action (Table~\ref{table:action}), and the type of hand contact (Table~\ref{table:hand}) are also presented.

% \begin{figure}[h]
%     \centering
% \includegraphics[width=\linewidth]{fig/fig1.png}
% \caption{Eight scanned objects in the~\model dataset, including a suitcase, a backpack, a stool, a container, a laundry basket, a chair, a box, and a bin.}
% \label{figure:obj}
% \end{figure}

\begin{table}[h]
    \small
    	\caption{Distribution of the dataset by the level of resistance. The data is categorized by the mass of removable weight used. Note, the masses of the objects themselves are not factored into these measurements.} 
	\centering
	\begin{tabular}[b]{c|cc}
		\hline
		{Mass} & {Minutes} & {$\%$}  \\
        \hline
        $\textrm{0 kg}$ & 47.1 & 33.3 \\
        $\textrm{5 kg}$ & 18.2 & 12.9  \\
        $\textrm{10 kg}$ & 21.4 & 15.1 \\
        $\textrm{15 kg}$ & 23.8 & 16.8  \\
        $\textrm{20 kg}$ & 7.9 & 5.6  \\        
        $\textrm{25 kg}$ & 8.9 & 6.2  \\
        $\textrm{$>$30 kg}$ & 14.2 & 10.1  \\ 
    	\hline
	\end{tabular}
    \label{table:weight}
\end{table}

\begin{table}[h]
    \small
        \caption{Distribution of the dataset by the type of action.} 
	\centering
	\begin{tabular}[b]{c|cc}
		\hline
		{Action Type} & {Minutes} & {$\%$}  \\
        \hline
        $\textrm{Carry}$ & 107.3 & 75.9 \\
        $\textrm{Push}$ & 17.4 & 12.3  \\
        $\textrm{Pull}$ & 16.7 & 11.8  \\
    	\hline
	\end{tabular}
        \label{table:action}
\end{table}
\begin{table}[h]
    \small
        \caption{Distribution of the dataset with different hand contact.} 
        \centering
        \begin{tabular}[b]{c|cc}
        \hline
        {Interaction Type} & {Minutes} & {$\%$}  \\
        \hline
        $\textrm{Right Hand}$ & 22.7 & 19.4 \\
        $\textrm{Left Hand}$ & 27.4 & 16.0  \\
        $\textrm{Both Hand}$ & 91.4 & 64.6  \\

    % $\textrm{Multi-Action Modes}$ & \xmark & \xmark & \cmark \\
        \hline
        \end{tabular}
        \label{table:hand}

\end{table}

% \begin{table}[h!]
%     \small
%     	\caption{Distribution of the dataset with different hand contact.} 
% 	\centering
% 	\begin{tabular}[b]{c|cc}
% 		\hline
% 		{Interaction Type} & {Minutes} & {$\%$}  \\
%         \hline
%         $\textrm{Right Hand}$ & 22.7 & 19.4 \\
%         $\textrm{Left Hand}$ & 27.4 & 16.0  \\
%         $\textrm{Both Hand}$ & 91.4 & 64.6  \\

%         % $\textrm{Multi-Action Modes}$ & \xmark & \xmark & \cmark \\
%     	\hline
% 	\end{tabular}
%     \label{table:hand}
% \end{table}

\myparagraph{Details on human tracking.} The first stage of our human tracking is to fit the SMPL parametric model~\cite{smpl} to the point clouds captured by the Kinect cameras. We segment humans in captured RGB images using Detectron V2 \cite{wu2019detectron2}. The resulting masks are then used to segment the human from the RGB data, before the the human point cloud is lifted in 3D. To initialize the SMPL pose, we employ FrankMocap \cite{rong2021frankmocap} from the images. Subsequently, instance-specific optimization techniques \cite{alldieck19cvpr} are applied to fit the SMPL model to the segmented human point cloud via ICP. For more precise fitting, we further derive the SMPL shape parameters of each subject from 3D scans using \cite{bhatnagar2020ipnet}. This stage produces the SMPL parameters fitted to the cameras, but they can be noisy and erroneous due to occlusion.\\
The second stage of our tracking is to refine the IMU-captured motion, which is smoother and more robust against occlusion. We synchronize the IMU-captured motion with the Kinect-fitted results from the previous stage, then perform an optimization to further refine the IMU-captured motion with the previously fitted results. The resulting motion is smooth and accurately captures the contact between the human and the object.

\section*{3\quad Architecture and Training Details}
\label{sec:training}

The motion synthesis network, \emph{MNet}, adopts a mixture-of-expert structure \cite{moe}. Both the gating network and the prediction networks consist of three-layer fully-connected networks, with hidden dimensions of 128 and 512, respectively. The model employs 8 experts and is trained for 150 epochs using an Adam optimizer. The initial learning rate is set at 1e-4, and a cosine learning rate scheduler gradually reduces it to 5e-6. A batch size of 32 is utilized, and the complete training process takes approximately 9 hours on an NVIDIA V100 GPU.

The contact prediction network, \emph{CNet} encodes the object geometry $\geom$ through a three-layer fully connected network of shape \{512, 512, 64\}, the resistance $\resistance$,  human joint positions $\poseposition$ and desired action $\goalact$ in a separate network with identical shape. The latent vector $\latent$ of the VAE is of size 6. The weight of the Kullback-Leibler divergence $\beta$ is 0.1. We use the Adam optimizer with a learning rate of 1e-3 and train CNet for 150 epochs. The full training of a subject-specific model takes approximately 10 minutes on an NVIDIA V100 GPU. 

\section*{4\quad Architecture and Training Details}
\label{sec:training}

The motion synthesis network, \emph{MNet}, adopts a mixture-of-expert structure \cite{moe}. Both the gating network and the prediction networks consist of three-layer fully-connected networks, with hidden dimensions of 128 and 512, respectively. The model employs 8 experts and is trained for 150 epochs using an Adam optimizer. The initial learning rate is set at 1e-4, and a cosine learning rate scheduler gradually reduces it to 5e-6. A batch size of 32 is utilized, and the complete training process takes approximately 9 hours on an NVIDIA V100 GPU.

The contact prediction network, \emph{CNet} encodes the object geometry $\geom$ through a three-layer fully connected network of shape \{512, 512, 64\}, the resistance $\resistance$,  human joint positions $\poseposition$ and desired action $\goalact$ in a separate network with identical shape. The latent vector $\latent$ of the VAE is of size 6. The weight of the Kullback-Leibler divergence $\beta$ is 0.1. We use the Adam optimizer with a learning rate of 1e-3 and train CNet for 150 epochs. The full training of a subject-specific model takes approximately 10 minutes on an NVIDIA V100 GPU. 

\end{document}